\documentclass[lettersize,journal]{IEEEtran}
\usepackage{amsmath,amsfonts}
\usepackage{algorithmic}
\usepackage{algorithm}
\usepackage{array}
\usepackage[caption=false,font=normalsize,labelfont=sf,textfont=sf]{subfig}
\usepackage{textcomp}
\usepackage{stfloats}
\usepackage{url}
\usepackage{verbatim}
\usepackage{graphicx}
\usepackage{cite}

\usepackage{multirow}
\usepackage{xcolor}
\usepackage{hyperref}

\usepackage[outdir=./]{epstopdf}

\begin{document}

\title{CNNs for JPEGs: A Study in Computational Cost}

\author{Samuel Felipe dos Santos, Nicu Sebe, and Jurandy Almeida
\thanks{Samuel Felipe dos Santos is with the Institute of Science and Technology, Federal
University of  S\~{a}o Paulo -- UNIFESP,
Brazil, e-mail: felipe.samuel@unifesp.br.}
\thanks{Jurandy Almeida is with the Dept. of Computing,
Federal University of S\~{a}o Carlos -- UFSCar,
Brazil, e-mail: jurandy.almeida@ufscar.br.}
\thanks{Nicu Sebe is with the Dept. of Information Engineering and Computer Science, 
University of Trento -- UniTn,
Italy, e-mail: niculae.sebe@unitn.it.}
\thanks{Manuscript received April 19, 2021; revised August 16, 2021.}}

\markboth{Journal of \LaTeX\ Class Files,~Vol.~14, No.~8, August~2021}%
{Shell \MakeLowercase{\textit{et al.}}: A Sample Article Using IEEEtran.cls for IEEE Journals}



\maketitle

\begin{abstract}
Convolutional neural networks (CNNs) have achieved astonishing advances over the past decade, defining state-of-the-art in several computer vision tasks.
CNNs are capable of learning robust representations of the data directly from the RGB pixels.
However, most image data are usually available in compressed format, from which the JPEG is the most widely used due to transmission and storage purposes demanding a preliminary decoding process that have a high computational load and memory usage. 
For this reason, deep learning methods capable of learning directly from the compressed domain have been gaining attention in recent years.
Those methods usually extract a frequency domain representation of the image, like DCT, by a partial decoding, and then make adaptation to typical CNNs architectures to work with them.
One limitation of these current works is that, in order to accommodate the frequency domain data, the modifications made to the original model increase significantly their amount of parameters and computational complexity.
On one hand, the methods have faster preprocessing, since the cost of fully decoding the images is avoided, but on the other hand,  the cost of passing the images though the model is increased, mitigating the possible upside of accelerating the method.
In this paper, we propose a further study of the computational cost of deep models designed for the frequency domain, evaluating the cost of decoding and passing the images through the network.
We also propose handcrafted and data-driven techniques for reducing the computational complexity and the number of parameters for these models in order to keep them similar to their RGB baselines, leading to efficient models with a better trade off between computational cost and accuracy.
\end{abstract}


\section{Introduction}
\IEEEPARstart{C}{onvolutional} neural networks (CNNs) have brought astonishing advances in computer vision, being used in several application domains, such as medical imaging, autonomous driving, road surveillance, and many others~\cite{ITSC_2019_Deguerre,li2019learning}.

However, in order to increase the performance of these methods, increasingly deeper architectures have been used, leading to models that consume a great amount o memory and demand a high computation cost for inference~\cite{liu2021group}.

For this reason, in spite of all the advances, the computational cost is still one of the main problems faced by deep learning architectures~\cite{ICCV_2019_Ehrlich}.
The significant amount of model parameters to be stored and the high GPU processing power needed for using such models can prevent their deployment in computationally limited devices, like mobile phones and embedded devices~\cite{ding2020prune}.
Therefore, specialized optimizations at both software and hardware levels are an imperative need for developing efficient and effective deep learning-based solutions~\cite{ISVLSI_2019_Marchisio}.
For storage and transmission purposes, most image data available are often stored in a compressed format, like JPEG, PNG and GIF~\cite{deguerre2021object}.
From these formats, JPEG has remained the most popular despite the advances in video compression and is considered a simple solution to store and transmit visual data~\cite{ehrlich2021analyzing}.
To use this data with a typical CNN, it would be required to decode it to obtain the RGB images used as input, a task demanding high memory and computational cost~\cite{ITSC_2019_Deguerre}.
And in edge devices with low computational power, like embedded systems, this decoding step can also became a bottleneck~\cite{wang2022fd}.
A possible alternative to alleviate this problem is to design CNNs capable of learning with DCT coefficients rather than RGB pixels~\cite{ITSC_2019_Deguerre,ICCV_2019_Ehrlich,deguerre2021object,ehrlich2021analyzing,wang2022fd,NIPS_2018_Gueguen,lo2019exploring,xu2020learning,ehrlich2020quantization,SIBGRAPI_2019_Santos,SIBGRAPI_2020_Santos,rajesh2019dct}.
The DCT is a representation of the data in the frequency domain that can be easily extracted by partial decoding, saving computational cost.
Frequency domain image processing can yield advantages like computational efficiency and spatial redundancy removal, being used successfully in many computer vision tasks, like image compression, image coding, signature verification, gender classification, face recognition, human gait recognition, lane detection, brain tumor classification, among others~\cite{tang2020facial,he2021fast,deshpande2021dct}.

For this study, we consider state-of-the-art CNNs recently proposed by Gueguen~et~al.~\cite{NIPS_2018_Gueguen}, which are modified versions of the ResNet-50 architecture~\cite{CVPR_2016_He}.
Despite the speed-up obtained by partially decoding compressed data, the architectural changes made to the model  
lead to a significant decrease computational efficiency.

To deal with this limitation, we propose an deeper study on the computational efficiency of preprocessing data and passing it trough the network for frequency domain models.
To alleviate the computational complexity and number of parameters, we extended the networks of Gueguen~et~al.~\cite{NIPS_2018_Gueguen} to include handcrafted and data-driven preprocessing techniques for reducing the amount of input channels that are fed to the network.

Extensive experiments were conducted, starting with ablation studies on a subset of ImageNet, analysing the effects of different frequency ranges, image resolution, JPEG quality and classification difficulty.
Then, we conduct experiments on the 
ImageNet dataset. 
%
The reported results indicate that by using our handcrafted and data-driven strategies to reduce the amount of input channels, we were able to reduce the computational complexity of such networks, making them faster than the RGB baseline. This way, it is possible to keep speed up by avoiding the decoding of images without increasing the computational cost of the model, leading to better trade offs between accuracy and computational efficiency.

Preliminary versions of this work were presented at IEEE International Conference on Image Processing (ICIP 2020)~\cite{ICIP_2020_Santos} and Iberoamerican Congress on Pattern Recognition (CIARP 2021)~\cite{santos2021less}, where we presented our handcrafted and data-driven approaches, respectively, to reduce the amount of input channels and computational complexity.
Here, we introduce several innovations. First, we present an in-depth review of deep learning methods that take advantage of the JPEG compressed domain. In addition, we discuss new strategies for reducing the amount of input channels that are fed to the network, testing the impact of not upsampling the chroma components of the DCT. Finally, we also include new experiments evaluating the performance of different models regarding their network inference speed in terms of frames per second (FPS), considering both the data preprocessing time and the network time.
The experiments we added were crucial for the discussions proposed in this paper. The inference time measurements provide further insight on the relevance of the decodification on the running time of the model and show that even if we disregard the gains in preprocessing time that were obtained by our strategies, they would still be considerably faster than the other compared models, a speed up that is crucial in environments with limited computational resources.

The remainder of this paper is organized as follows.
Section~\ref{sec:compression}  briefly reviews the JPEG image compression algorithms.
Section~\ref{sec:related} discusses related work.
Section~\ref{sec:approach} describes our approach to reduce the computational cost for decoding images by training CNNs directly on the compressed data.
Section~\ref{sec:results} presents the experimental setup and reports our results.
Finally, we offer our conclusions and directions for future work in Section~\ref{sec:conclusions}.

\section{JPEG Compression}
\label{sec:compression}
\begin{figure}[!htb]
    \centering
    \includegraphics[width=0.9\columnwidth]{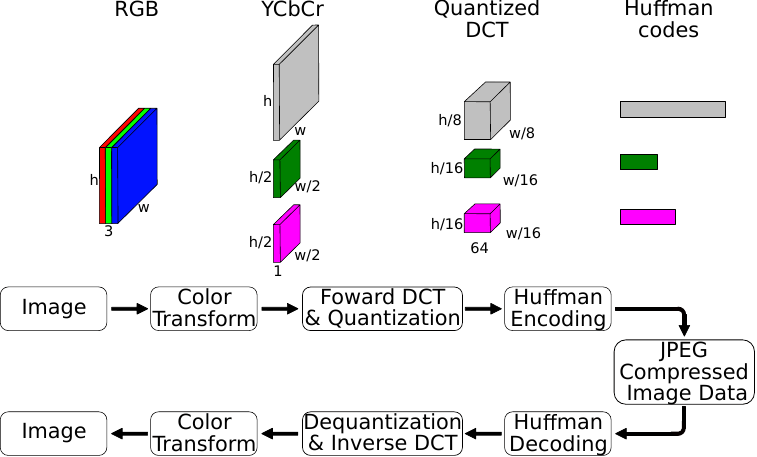}
    \caption{JPEG compression and decompression process~\cite{NIPS_2018_Gueguen}.}
    \label{fig:jpeg}
\end{figure}

The JPEG standard (ISO/IEC 10918) was created in 1992 and is currently the most widely-used image coding technology for lossy compression of digital images. 
The basic steps of the JPEG compression algorithm are presented in Figure~\ref{fig:jpeg}.

Initially, the representation of the colors in the image is converted from RGB to YCbCr, which is composed of one luminance component (Y), representing the brightness, and two chrominance components, Cb and Cr, representing the color. 
Also, acording to the subsampling scheme selected, the Cb and Cr components can be down-sampled horizontally and vertically, since human vision is more sensitive to brightness details than to color details.
Some of the most commonly used subsampling schemes are 4:4:4, where there is no chroma subsampling at all, 4:2:2, where the spatial resolution of the Cb and Cr channels are reduced by a factor of 2 horizontally, and  4:2:0, where the resolution is reduced in half both horizontally and vertically.
Then, each of the three components are partitioned into blocks of 8$\times$8 pixels and 128 is subtracted from all the pixel values. 
Next, each block is converted to a frequency domain representation by the forward discrete cosine transform~(DCT).
The result is an 8$\times$8 block of frequency coefficient values, each corresponding to the respective DCT basis functions, with the zero-frequency coefficient (DC term) in the upper left and increasing in frequency to the right and down.
The amplitudes of the frequency coefficients are quantized by dividing each coefficient by a respective quantization value defined in quantization tables, followed by rounding the result to the the nearest integer.
High-frequency coefficients are approximated more coarsely than low-frequency coefficients, as human vision is fairly good at seeing small variations in color or brightness over large areas, but fails to distinguish the exact strength of high-frequency brightness variations.
The quality setting of the encoder affects the extent to which the resolution of each frequency component is reduced.
If an excessively low-quality setting is used, most high-frequency coefficients are reduced to zero and thus discarded altogether.
To improve the compression ratio, the quantized blocks are arranged into a zig-zag order and then coded by the run-length encoding (RLE) algorithm.
Finally, the resulting data for all 8$\times$8 blocks are further compressed with a lossless algorithm, a variant of Huffman encoding. For decompression, inverse transforms of the same steps are applied in reverse order. If the DCT computation is performed with sufficiently high precision, quantization and subsampling are the only lossy operations as the others are lossless, therefore they are reversible.

\section{Related Work}
\label{sec:related}


In this section we present a literature review of deep learning methods that use DCT coefficients.
We divided such works into two categories according to how they are related to our approach:
(1) works that aim to improve performance of CNNs by using information from the frequency domain, like DCT coefficients and (2) works that focus on deep model designed to process data directly from the frequency domain in the form of DCT coefficients.
In the first category, we can highlight the following works~\cite{ehrlich2021analyzing,ehrlich2020quantization,temburwar2022deep}.

Ehrlich~et~al.~\cite{ehrlich2020quantization} proposed a single model that is capable of dealing with variable JPEG qualities, called Quantization Guided JPEG Artifact Correction~(QGAC).
This is achieved by parameterizing the CNN with the quantization matrix of the JPEG images with the Convolutional Filter Manifold~(CFM), that is a variation of the Filter Manifolds, and uses a CNN to output a convolutional kernel thought the quantization matrix, adapting the weights and bias according to JPEG quality level.
Also proposed by Ehrlich~et~al.~\cite{ehrlich2021analyzing} the Task-Target Artifact Correction is a method that mitigate the performance penalty of using JPEG images with not ideal compression qualities.
Temburwar~et~al.\cite{temburwar2022deep} tackled another task with DCT CNNs, the Content-based image retrieval (CBIR), avoiding the cost of decoding.

On the second category, we highlight the following works~\cite{ITSC_2019_Deguerre,ICCV_2019_Ehrlich,deguerre2021object,wang2022fd,NIPS_2018_Gueguen,lo2019exploring,xu2020learning} modify CNNs specifically to deal with DCT coefficients.

Ehrlich~and~Davis~\cite{ICCV_2019_Ehrlich} reformulated the ResNet architecture to perform its operations directly on the JPEG compressed domain.
Lossless operations like convolution, batch normalization and others that are linear where adapted to operate on the JPEG compressed domain, while the non-linear one, like ReLU, were approximated.
In a different direction,  Lo~et~al.~\cite{lo2019exploring} explore the task of semantic segmentation directly on the DCT representation  of the images.
The proposed method uses the Frequency Component Rearrangement (FCR) technique to code the relationship between the DCT coefficients in a new dimension, feeding them to the proposed DCT-EDANet, a version of the EDANet~\cite{lo2019efficient} architecture modified to operate on the frequency domain.
Xu~et~al.~\cite{xu2020learning} proposed a method that can be applied to deep models with few modifications to the architecture in order to use information from higher resolution images, mitigating the loss of salient information caused by downsampling.
The method consists of extracting the DCT coefficients of the image in order to obtain a representation on frequency domain, followed by a training process where the network learns to dynamically select only a subset of the most relevant frequencies jointly with the task.
To accelerate the training and inference speed, Gueguen~et~al.~\cite{NIPS_2018_Gueguen} proposed different architectural modifications to apply to the ResNet-50 network~\cite{CVPR_2016_He} in order to accommodate DCT coefficients.
These coefficients can be obtained by partial decoding, thus saving the high computational load and memory usage in full decoding the JPEG images.
The modifications made to accommodate the DCT consisted of skipping the first stage of the ResNet-50 network and alterations to the early blocks of the second and third stages to mimic the increase in receptive fields and stride of the baseline RGB model, calling this model Receptive Field Aware (RFA).
In order to deal with the lower resolution of the Cb and Cr components, multiple methods were proposed.
The Upsampling-RFA upsample the Cb and Cr components to match the resolution of the Y component and the Deconvolution-RFA do the same process, but with a deconvolution layer instead. On the Late Concat RFA (LC-RFA), the Y component goes thought the modified versions of the second and third stages of the network, while the Cb and Cr components goes thought a separate convolution block.
Then, all components are concatenated and fed to the fourth stage.
Gueguen~et~al.~\cite{NIPS_2018_Gueguen} also proposed the LC-RFA-Thinner, a versions of the LC-RFA with altered number of channels on the first three convolution blocks that the Y component goes though, respectively, from 1024, 512, 512 channels, to 384, 384, 768.
%
Similarly, Deguerre~et~al.~\cite{ITSC_2019_Deguerre} proposed a fast object detection method which takes advantage of the JPEG compressed domain.
For this, the Single Shot MultiBox Detector (SSD)~\cite{ECCV_2016_Liu} architecture was adapted to accommodate block-wise DCT coefficients as input. 
%
Deguerre~et~al.~\cite{deguerre2021object} also extended their work by
using the LC-RFA, Deconvolution-RFA and LC-RFA-Thinner networks from Gueguen~et~al.~\cite{NIPS_2018_Gueguen} as backbone to the SSD architecture, that originally used the VGG-16, naming them SSD300.
Wang~et~al.~\cite{wang2022fd} also followed a similar strategy, skipping the first stage of the Tiny-YOLO-v3, ResNet-18 and SqueezeNet networks, feeding then with DCT coefficients, but used the models on the FPGA ZYNQ and proposed optimization strategies for the decoding step on the device. The results showed that on embedded devices like these, that are commonly used for CNN acceleration, the decoding step can be a bottleneck for execution time, this way, using CNNs designed for frequency domain data can have even greater speed ups, being up to 4.29 times faster.

There are also other recent works that use information from the frequency domain but where not included in more details, since they have some key aspect that distance it from the scope of our work.
For instance,
Chen~et~al.~\cite{chen2018learning} uses Wavelet-like Auto-Encoder (WAE) to decompose the original image into two sub-images with lower resolution, reducing the computational cost;
Chamain~et~al.~\cite{chamain2019faster} proposed a model to avoid decompression for JPEG2000 images, designing a CNN that works with Discrete Wavelet Transformed (DWT) coefficients;
Li~et~al.~\cite{li2020learning} performed artefact removal on JPEG images with multiple quality settings but used only quantization tables from the JPEG compressed data;
Chen~et~al.~\cite{chen2021compressed} designed a network for video super-resolution using partition maps and motion vectors;
Tang~et~al.~\cite{tang2020facial} used a network designed for frequency data for facial expression recognition
and Qin~et~al.~\cite{qin2021fcanet} proposed the FcaNet, a model that uses frequency channel attention, weighting the importance of each input channel.

For the category of CNN designed specifically for DCT coefficients, some works have been proposed recently, as the ones mentioned so far, but in general they have only focused on improving the effectiveness of a given model.
When analysing models designed for DCT coefficients, it is also very important to take into account their efficiency, since one of their main advantages is the speed up obtained by partially decoding the images, but in order to increase accuracy, many works end up increasing the complexity of the model, losing this advantage.
In spite of all the advances in this nascent area, a more comprehensive study considering efficiency aspects, such as computational complexity and number of parameters, is still missing.
%
This paper aims to fill such a gap. Here, we present a comprehensive study of DCT based models that considers not only their effectiveness but also their efficiency, such as computational complexity, number of parameters, and inference time. In addition, we propose different strategies to reduce the computational complexity and number of parameters of such models, in order to keep them similar to their RGB baselines.

\section{Learning from the Compressed Domain}
\label{sec:approach}

The DCT computation of a 8$\times$8 pixel block requires 1920 floating point operations (FLOPs)~\cite{BOOK_2007_Hanzo}. 
Although it seems negligible, computer vision tasks usually involve the processing of a great amount of images, each one containing many pixel blocks, therefore the total computational cost may be significant. 
For example, on the ImageNet dataset,  considering the resolution of 224$\times$224 as input size of the network, it would be necessary 150.52 GFlops for decoding the 100,000 images of the test set, a considerable amount, and 1,806.25 GFlops for the 1,2 million images of the training set. 
Those are significant values  and reflect on the overall speed of the model.

Roughly speaking, the DCT can be seen as a convolution with a specific filter size of  $8 \times 8$, stride of  $8 \times 8$, one input channel, 64 output channels, and specific, non-learned orthonormal filters. 
As both the filter size and stride are equal to 8, spatial information of adjacent blocks do not overlap. 
In theory, a standard convolutional layer may learn to behave like a DCT, but in practice, this is not trivial, as the learned bases may be undercomplete, complete but not orthogonal, or overcomplete.
In spite of that, the use of DCT weights as input for a CNN is feasible, since they can be seen as the outputs of a convolution layer with frozen weights initialized from the DCT filters~\cite{NIPS_2018_Gueguen}.
Motivated by the aforesaid observations, we examine ways of integrating frequency domain information into CNNs.
To present date, little work has been done to exploit the DCT representation widely used in compressed data as input for neural networks~\cite{NIPS_2018_Gueguen}.

The starting point for our proposal were the models from Gueguen~et~al.~\cite{NIPS_2018_Gueguen}, which are adapted to facilitate the learning with DCT coefficients rather than RGB pixels.
All models proposed by Gueguen~et~al.~\cite{NIPS_2018_Gueguen} had similar accuracy, being the main difference the way they deal with the lower resolution of the chroma components. 
Initially, we decided to focus on the Upsampling-RFA, since it is the model with the least amount of changes to the original ResNet-50.
Later, in order to attempt to reduce the preprocessing cost even further, we adapted our strategies to the LC-RFA, that have a greater amount of architectural changes in comparison with the ResNet-50, but do not upsample the chroma components.

On the Upsampling-RFA, DCT coefficients obtained from the Cb and Cr components are upsampled in order to match the resolution of the Y component. 
Then, the three components are concatenated channel-wise, passed through a batch normalization layer, and fed to the convolution block of the second stage of the ResNet-50 network. 
The second and third stages of the ResNet-50 network were changed to accommodate the amount of input channels and to ensure that the number of output channels at the end of these stages is the same of the original ResNet-50 network.
Due to the smaller spatial resolution of the DCT inputs, early blocks of the second stage of the ResNet-50 network were altered to have a smoother increase of their receptive fields and, for this reason, their strides were decreased in order to mimic the increase in size of the receptive fields in the original ResNet-50 network.

\begin{figure}[!h]

    \begin{minipage}[b]{1.0\linewidth}
      \centering
      \centerline{\includegraphics[width=\textwidth]{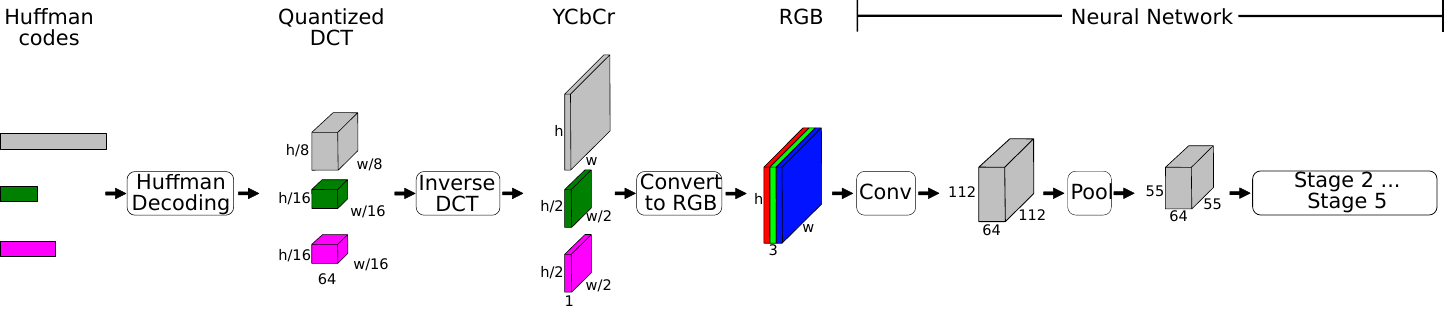}}
      \centerline{(a) Original ResNet-50 network}\medskip
    \end{minipage}
    \begin{minipage}[b]{1.0\linewidth}
      \centering
      \centerline{\includegraphics[width=\textwidth]{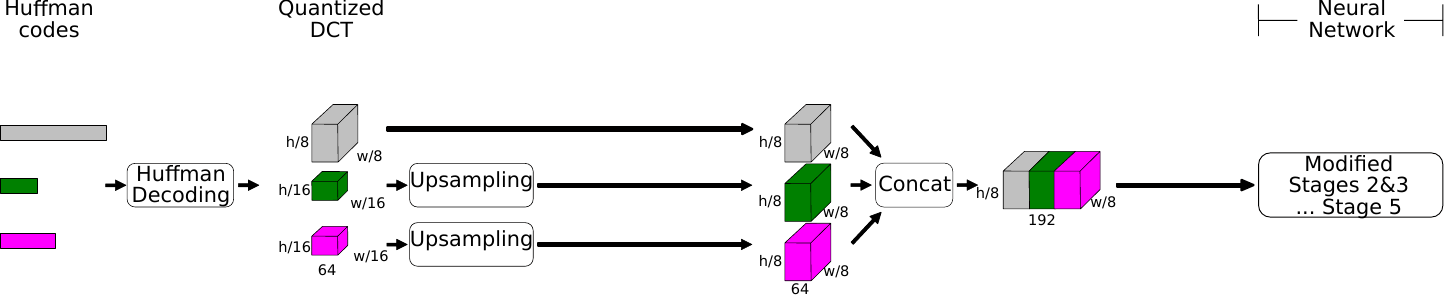}}
      \centerline{(b) Upsampling-RFA}\medskip
    \end{minipage}
    \hfill
    \begin{minipage}[b]{1.0\linewidth}
      \centering
      \centerline{\includegraphics[width=\textwidth]{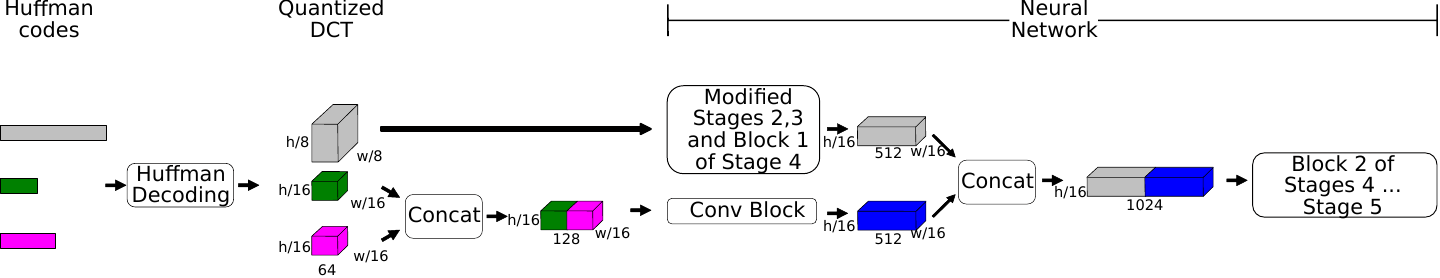}}
      \centerline{(c) LC-RFA}\medskip
    \end{minipage}
    \hfill
    \begin{minipage}[b]{1.0\linewidth}
      \centering
      \centerline{\includegraphics[width=\textwidth]{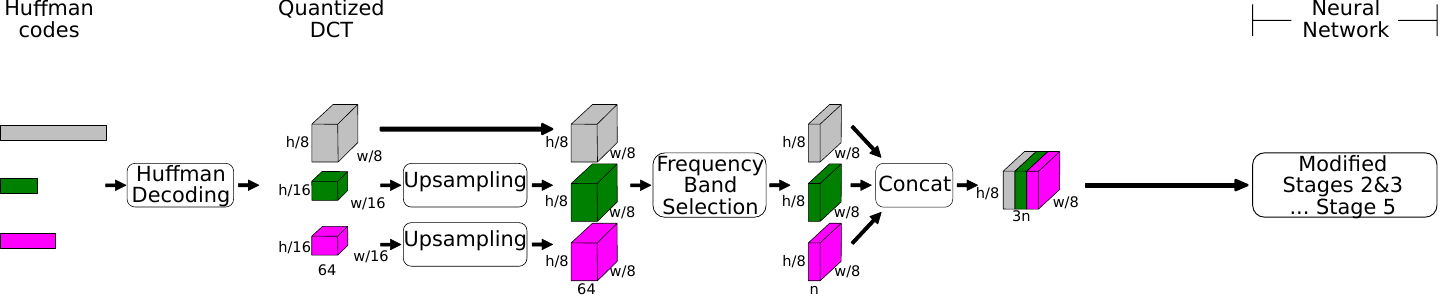}}
      \centerline{(d) Upsampling-RFA with handcrafted channel reduction (ours)}\medskip
    \end{minipage}
    
    \hfill
    \begin{minipage}[b]{1.0\linewidth}
      \centering
      \centerline{\includegraphics[width=\textwidth]{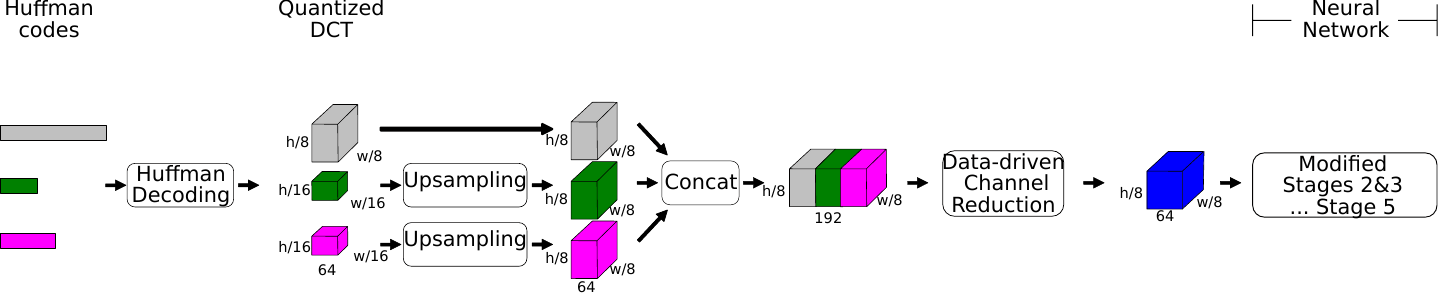}}
      \centerline{(e) Upsampling-RFA with data-driven channel reduction (ours)}\medskip
    \end{minipage}
    \hfill

    \begin{minipage}[b]{1.0\linewidth}
      \centering
      \centerline{\includegraphics[width=\textwidth]{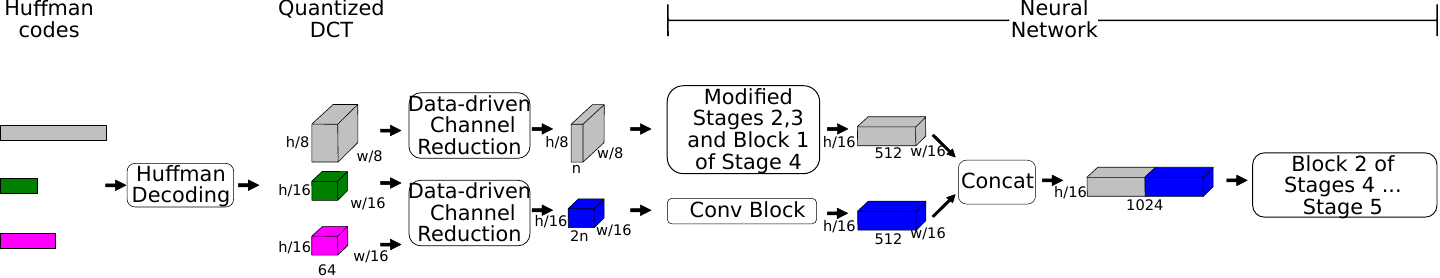}}
      \centerline{(f) LC-RFA with data-driven channel reduction (ours)}\medskip
    \end{minipage}    
    \hfill
    
    \begin{minipage}[b]{1.0\linewidth}
      \centering
      \centerline{\includegraphics[width=\textwidth]{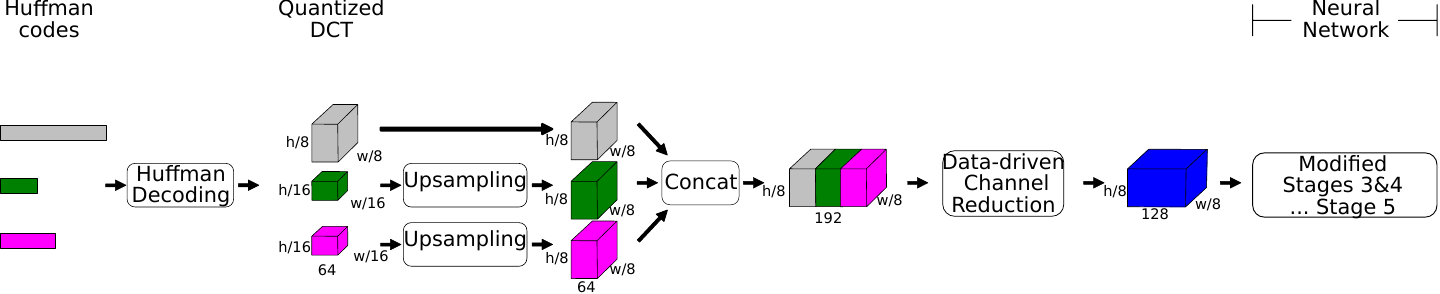}}
      \centerline{(g) Upsampling-RFA with reduced amount of layers (ours)}\medskip
    \end{minipage}
    \hfill

    \begin{minipage}[b]{1.0\linewidth}
      \centering
      \centerline{\includegraphics[width=\textwidth]{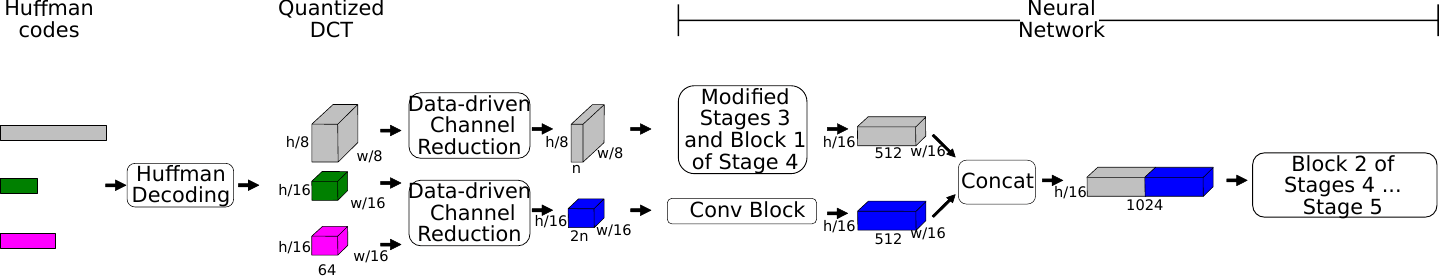}}
      \centerline{(h) LC-RFA with reduced amount of layers (ours)}\medskip
    \end{minipage}
    \caption{Illustrations of (a) the original ResNet-50 network~\cite{CVPR_2016_He}, (b) the Upsampling-RFA and (c) the LC-RFA network proposed by Gueguen~et~al.~\cite{NIPS_2018_Gueguen},
    our improved versions from the Upsamling-RFA with (d) handcrafted and (e) data-driven techniques to reduce the amount of channels, (f) our improved versions from the LC-RFA with data-driven techniques, and
    our method to reduce the amount of layers with data-driven techniques on the (g) Upsampling-RFA and on the (b) LC-RFA.}
    \label{fig:FBS}
\end{figure}

\clearpage

On the LC-RFA, the Y component goes thought modified versions of stages 2, 3, and the first convolutions block of stage 4, all of them with the same stride reduction and increased amount of filters as the Upsampling-RFA, meanwhile, the Cb and Cr components goes thought a separate convolution block.
They are then joined and fed to the second convolutional block of stage 4 of the network.
These changes in the ResNet-50 network raised its computation complexity and number of parameters.
Although there was a the speed-up by partially decoding the images, there was also an increase in the computational cost for passing them through the network once they are loaded in memory.
Also, the increase in the amount of parameters indicates that in order to reach a similar classification accuracy from the RGB model, the Upsampling-RFA and LC-RFA needed to use a more robust model than the original. 
In this way, the speed-up obtained by partially decoding the images is outweighed by the increase in the computational cost for passing them through the network once they are loaded in memory.

Motivated by the above observations, we examine different ways of dealing with those issues.
Our objective was to obtain a model that works directly on DCT coefficients, but also have a similar amount of parameters and computational complexity from the original ResNet-50 for RGB images.
Reducing preprocessing time, computational complexity and number of parameters, but keeping similar classification performance is
paramount on devices with limited resources, like mobile devices and embedded systems, facilitating the use of deep models on a wider array of different application and environments.
A comparison among the original ResNet-50 network~\cite{CVPR_2016_He}, the Upsampling-RFA, the LC-RFA~\cite{NIPS_2018_Gueguen}, and our proposed techniques is presented in Figure~\ref{fig:FBS}.
Initially, in Section~\ref{sec:fbs}, we proposed a handcrafted technique named Frequency Band Selection (FBS), where we manually select a subset of the most relevant DCT coefficients before feeding them to Upsampling-RFA;
Then in Section~\ref{sec:channels}, we tested data-driven techniques to reduce the amount of input channel and in Section~\ref{sec:lcrfa-ccpp} we studied the effects of avoiding the upsampling of the chroma components;
Finally, in Section~\ref{sec:layers} we explored the effects of reducing the amount of layers of the CNN combined with data-driven techniques.

\subsection{Reduce Input channels with a Handcrafted technique}
\label{sec:fbs}

To alleviate the computational complexity and the number of parameters, of the Upsampling-RFA~\cite{NIPS_2018_Gueguen}, we proposed to use a Frequency Band Selection (FBS) technique to select the most relevant DCT coefficients before feeding them to the network. 
Since higher frequency information has little visual effect on the image, we retain the $n$ lowest frequency coefficients and analyze how they impact the classification accuracy of the network.
For that, the second stage of the ResNet-50 network was changed to accommodate the amount of retained coefficients, i.e., $3 n$ input channels.
In this way, when FBS is set to $n=64$, the network architecture is the same as Upsampling-RFA~\cite{NIPS_2018_Gueguen} and the number of output channels at the end of the second stage is equals to 256, while for $n=16$, the number of output channels is the same amount of the original ResNet network~\cite{CVPR_2016_He}, i.e., 64.

\subsection{Reduce Input channels with with Data-driven Techniques}
\label{sec:channels}

The results obtained by the FBS technique proposed in Section~\ref{sec:fbs} indicates that reducing the number of channels in early stages of the network can be effective in reducing the computational costs of the network.
For this reason, we decided to explore smarter strategies to make this reduction.

First, we reduce the number of input channels of the second stage to 64 but we kept the decreased strides at its early blocks, as proposed by Gueguen~et~al.~\cite{NIPS_2018_Gueguen}. 
To accommodate this amount of input channels, we change number of output channels of the second and third stages are changed to be the same as the original ResNet-50.
Then, we evaluate different strategies to reduce the number of DCT inputs from 192 (i.e., 3 color components $\times$ 64 DCT coefficients) to 64. 

Unlike the FBS, where the DCT inputs are discarded by hand potentially losing image information, these techniques take advantage of all DCT inputs and learn how to combine them in a data-driven fashion.
For this, we evaluate three different approaches: (1) a linear projection~(LP) of the DCT inputs (Section~\ref{sec:lp}), (2) a local attention~(LA) mechanism (Section~\ref{sec:la}), and (3) a cross channel parametric pooling~(CCPP) (Section~\ref{sec:nin}).

\subsubsection{Linear projection (LP)}
\label{sec:lp}

The ResNet-50~\cite{CVPR_2016_He} have residual learning applied to every block of few stacked layers, given by Equation~\ref{eq:lp1}, where $F()$ is the residual mapping to be learned by the $i$-th block of stacked layers, $W_i$ are its parameters, $x$ are the input data, and $y$ are the output feature maps.
\begin{equation}
    \label{eq:lp1}
    y = F ( x, {W_i} ) + x 
\end{equation}

The $F() + x$ operation is executed by a shortcut connection and a element-wise addition, but their dimensions must be equal. When they are not, a $W_s$ linear projection can be applied in order to match the dimension. 
As can be seen in Equation~\ref{eq:lp2}, assuming that $x$ have $n$ input features maps and $W_s$ is a weight matrix of size $m \times n$, the product $W_s \cdot x$ will output in $m$ feature maps, where each one is a linear combination of all the $n$ inputs from $x$.
\begin{equation}
    \label{eq:lp2}
    y = F( x, {W_i} ) + W_s \cdot x
\end{equation}

We apply this linear projection to reduce the number of channels from 192 to 64 of the convolution block of the second stage. 
In this way, we consider the DCT inputs as a whole regardless the importance of each of their frequencies to the image content. 
Also, the skewness or kurtosis (shape) of their distribution is preserved by the linear transformation.

\subsubsection{Local attention (LA)}
\label{sec:la}

Visual attention is the cognitive process where, given a natural scene, the most significant visual information is selected while other redundant information is filtered out, being an important mechanism for the human visual system~\cite{fang2012saliency}.
Attention mechanics for CNNs can have many types of variants, like spatial, channel and self-attention, according to the dimensions of the input they focus.
In the case of channel attention, importance weights are learned and attached to the input channels~\cite{qin2021fcanet}. 

The local attention proposed by Luong~et al.~\cite{luong2015effective} is a soft attention mechanism used on the machine translation task to analyze a word with a small context window of adjacent words, learning attention maps which focus on relevant parts of the input information.
We adapt this mechanism to be used as channel attention for the DCT inputs in order to reduce the number of channels from 192 to 64. This is performed according to Equation~\ref{eq:la}, where $x$ is an input with $n$ features maps, $r$ is its reshaped version partitioning it into $m$ groups of $\frac{n}{m}$ channels, $W$ is a weight matrix of size $m \times (\frac{n}{m})$, $y$ is an output with $m$ feature maps, and $\odot$ is the Hadamard product.
\begin{align}
    \label{eq:la}
    r &= reshape\left( x, \left[ m, \frac{n}{m} \right] \right) \\
    s &= W \odot r \nonumber \\
    a_i &= softmax( s_i ), \forall i \in \{1 \dots m\} \nonumber \\
    y_i &= a_{i} \cdot r_{i}, \forall i \in \{1 \dots m\} \nonumber
\end{align}

First, the input $x$ is split into $m$ partitions $r = \{r_1, \dots, r_m\}$ with $\frac{n}{m}$ features maps.
Then, alignment scores $s$ are obtained by computing the Hadamard product between $W$ and $r$. 
For each partition $i \in \{1 \dots m\}$, alignment scores $s_i$ are normalized by applying the softmax function, producing attention maps $a_i$ which are used to amplify or attenuate the focus of the distribution of the input data $r_i$. 
Therefore, the feature map $y_i$ outputted for the $i$-th partition is a linear combination of adjacent channels. 
In this way, we preserve information of the DCT spectrum for the entire range of frequencies.

\subsubsection{Cross Channel Parametric Pooling (CCPP)}
\label{sec:nin}

In a cross channel parametric pooling layer, a weighted linear recombination of the input features maps is performed and then passed though a rectifier linear unit (ReLu)~\cite{lin2013network}.
Min~Lin~et~al.~\cite{lin2013network} proposed to use a cascade of such layers to replace the usual convolution layer of a CNN, since they have enhanced local modeling and the capability of being stacked over each other. 
Formally, a cascaded cross channel parametric pooling is performed according to Equation~\ref{eq:nin}~\cite{lin2013network}, where $f_{i,j,k}^l$ stands for the output of the $l$-th layer, $x_{i,j}$ is the input patch centered at the pixel $(i, j)$, $k$ is used to index the feature maps, $W_{l,k}$ and $b_{l,k}$ are, respectively, weights and biases of the $l$-th layer for the $k$-th filter, and $N$ is the number of layers.     
\begin{align}
    \label{eq:nin}
    f_{i,j,k}^1 &= max\left( 0, W_{1,k}^T \cdot x_{i,j} + b_{1,k} \right) \\
    &\vdots \nonumber \\
    f_{i,j,k}^N &= max\left( 0, W_{N,k}^T \cdot f_{i,j}^{N-1} + b_{N,k} \right)\nonumber
\end{align}

The cross channel parametric pooling is equivalent to a convolutional layer with a kernel size of $1\times1$~\cite{lin2013network}, which is also know as a pointwise convolution~\cite{chollet2017xception}, being capable of projecting the input feature maps into a new channel space, increasing or decreasing the amount of channels.

We used a cross channel parametric pooling layer to reduce the number of channels from 192 to 64.
Similar to the linear projection, the individual importance of each DCT coefficient for the image content is also not taken into account.
On the other hand, the non-linear properties of the ReLU activation encourages the model to learn sparse feature maps, making it less prone to overfitting.

\subsection{Avoiding the Chroma Upsampling}
\label{sec:lcrfa-ccpp}

All our strategies proposed so far are modifications made to the Upsampling-RFA. We decided to start our experiments with this model because, between the strategies proposed by Gueguen~et~al.~\cite{NIPS_2018_Gueguen}, this was the one with the least amount of architectural changes in comparison to the ResNet-50, but our proposed strategies can also be applied to other network for DCT coefficients with ease.
With this motivation, we adapted our strategies to the LC-RFA, a model with more complex architectural changes than the Upsampling-RFA, but one that do not apply upsample chroma components of the DCT, reducing even further the preprocessing cost.

The original LC-RFA have initially two streams of data, one for the Y channel of the DCT and one for the concatenated Cb and Cr channels. The output of both streams are latter concatenated and feed to the second convolutional block from stage 4.  
The Y stream have modified versions of stages 2, 3 and the first convolutional block from stage 4 of the original ResNet-50.
While the CbCr stream have a convolutional block similar to the first block from stage 4 but with kernel size of 1.
The amount of channels and stride from the layers are the same from the Upsampling-RFA, with the only difference being the first block from stage 4 in both streams, that have the amount of input and output channels halved, so when concatenated, they have the expected input size of the second block from stage 4 of the model.
In our previous experiments with data-driven methods in the Upsampling-RFA, we kept the stride reduction on early layers, but we used the same amount amount of convolutional channels as the original ResNet-50. 
Then, we applied different strategies to reduce amount of channels from 192 to 64, the expected input shape for the second stage of the original ResNet-50, since that is the input size from the original third stage.

For our proposed models based on the LC-RFA, we kept the same strategy, applying the stride reduction from Gueguen~et~al.~\cite{NIPS_2018_Gueguen}, but using the same amount of convolutional channels from the original ResNet-50.
But if we keep the input size of Y stream the same to the respective layers in the original ResNet-50, it would be 64, and for the CbCr stream, if we use half the input size of the respective layer, since the LC-RFA halves the input size and amount filters from this layer in order to account to the fact that it is repeated in both streams, it would be 128.
Those are the values already supplied in DCT coefficients, 64 Y channels and 64 channels for Cb and Cr each, totaling 128. For this motive, we decided to use our CCPP strategy to reduce the amount of input channels from each stream, allowing us to reduce the computational complexity even further.

This way, in our first experiment, two modules of one of our data-driven strategies were added, one for Y stream and one for the CbCr stream, they reduce the amount of channels to 32 Y and 64 CbCr channels or 16 Y and 32 CbCr channels. The amount of CbCr channels are the double from Y, since they together (Cb and Cr) have double the amount of coefficients.

\subsection{Reducing the Number of Layers }
\label{sec:layers}

Both the Upsampling-RFA and LC-RFA~\cite{NIPS_2018_Gueguen} skips the first stage of the original ResNet-50, feeding the DCT coefficients to the second stage, which is modified to accommodate the amount of input channels.
In order to reduce the complexity and amount of parameters from the network even further, we analyze the effects of skipping the second, third, and fourth stages of from the Upsampling-RFA, but maintaining the stride reduction proposed by Gueguen~et~al.~\cite{NIPS_2018_Gueguen} at the early blocks of the initial stage in which the DCT coefficients are provided as input.
Different from Gueguen~et~al.~\cite{NIPS_2018_Gueguen}, we do not increase the number of input channels at the initial stages, since it would lead to a great increase on the computational complexity of the network.
Instead, we keep them the same as the original ResNet-50, whose the number of input channels in the second, third, fourth, and fifth stages are 64, 128, 256, and 512, respectively.
To accommodate such amount of channels in the Upsampling-RFA, the data-driven strategies presented in the previous section were used to decrease or increase the DCT inputs from 192 (i.e., 3 color components $\times$ 64 DCT coefficients) to the amount of input channels of the initial stage in which they are provided as input.
Notice that the number of DCT coefficients is close to the number of input channels of the third stage, requiring a less drastic reduction than the one needed to feed them on the second stage.
On the other hand, the expected inputs for the fourth and fifth stages have a greater amount of channels than the DCT inputs and, for this reason, they need to be scaled up, however preserving the salient features as the original data.
Since our results on the Upsampling-RFA were promising, we also tested skipping the second stage on our LC-RFA models with data-driven techniques to reduce the amount of input channels.

\section{Experimental Protocol}
\label{sec:protocol}

All the experiments were executed on a machine equipped with a processor Intel Core i7 6850K 3.6 GHz, 64 GBytes of DDR4-memory, and 3 NVIDIA Titan Xp GPUs.
The machine runs Ubuntu 18.04 LTS (kernel 5.0.0) and the ext4 file system.
The models were implemented using torch 1.2.0, and the experiments were execute on the ImageNet dataset and a subset of it.
Since we did not had access to the original implementation, of Gueguen~et~al.~\cite{NIPS_2018_Gueguen} models, we re-implemented them following the specifications available and trained them from scratch.
In order to make a fair comparison, we present the results of our networks trained from scratch, as well as the results obtained from models trained from other works, showing that the variance present in the training of the network, and small variations in training procedures can generate variance in the results obtained by these models.

In order to compare our approach with the state-of-the-art, we selected models with similar alterations to the network architecture as our methods, attempting to reduce the amount of channels of the networks designed for DCT coefficients to make them more similar to the the original ResNet-50 for RGB images.
For this reason, we did not include models like the ones based on the LC-RFA Thinner~\cite{NIPS_2018_Gueguen}, since they have a smaller amount of channels than the original ResNet-50, although this modifications could be applied to our methods.

Experiments were performed using the ResNet-50, Upsampling-RFA, Upsampling-RFA + FBS, Upsampling-RFA + LP, Upsampling-RFA + LA, Upsampling-RFA + CCPP, Upsampling-RFA + CCPP skipping the 1$^{st}$ and 2$^{nd}$ stages, LC-RFA, LC-RFA + CCPP and LC-RFA + CCPP skipping stages 1 and 2, batch size of 128, initial learning rate of 0.05, momentum of 0.9, and were trained for 120 epochs, reducing the learning rate by a factor of 10 every 30 epochs. 
All images were resized so the smallest side is 256 and the crop size is 224$\times$224.
Data augmentation with random crop and horizontal flip was used on the training phase, while on test, only a center crop was used.

Initially, we conducted ablation studies with our handcrafted approach on a subset of the ImageNet~\cite{IJCV_2015_Russakovsky} dataset\footnote{The script for generating the ImageNet subset is available at \url{https://github.com/dusty-nv/jetson-inference/blob/master/tools/imagenet-subset.sh} (As of January, 2023)}.
For that, 211 of the 1000 classes were taken from the ILSVRC12 dataset and then grouped into 12 classes, namely: ball, bear, bike, bird, bottle, cat, dog, fish, fruit, turtle, vehicle, sign.
This subset is composed of 268,773 images and was split into a training set of 215,018 (80\%) images and a test set of 53,755 (20\%) images.
Two classification difficulties were tested on this subset, fine-grained, where 211 out of 1,000 classes from the ILSVRC12 dataset were used, and coarse-grained, where these classes are grouped into 12 categories.
Finally, We evaluated the classification accuracy of all our propose methods on the entire ImageNet dataset.

To compare the performance of the network on the image classification task, we used the accuracy metric.
In order to compare the efficiency of the networks, we used the amount of trainable parameters, the number of floating point operations (measured in GFLOPs), inference time and the frames per second (FPS).
For measuring inference time, we used a single Titan XP GPU and followed Deguerre~et~al.~\cite{deguerre2021object} by doing 10 runs of 200 predictions with batch size of 8.
The final inference times are the average over all runs.
Following Gueguen~et~al.~\cite{NIPS_2018_Gueguen}, we also loaded the data into the memory previously to the time measurements, but differently than these works, we loaded the JPEG compressed data, allowing us to analyze the effects of partially or fully decoding the images.   
We divided inference time on data preprocessing time and the time for passing the data through the network.
Data preprocessing time includes the time for fully or partially decoding the images, applying the necessary operations and loading them into the GPU memory.
All images were center cropped previously to the execution of these experiment.
We use a libjpeg-turbo based implementation for the partial and full decoding of the JPEG images.
In most of our experiments, FPS is calculated over the average network inference time, since different strategies can be used to speed up the decompression, like GPU acceleration, and most of other work report results this way, like Deguerre~et~al.~\cite{deguerre2021object}, except for the ones in Section~\ref{sec:time}, where we also show the FPS based on the total time including the preprocessing time, allowing us to see the impact of the decoding step.
FPS from other works are also reported, but are not directly comparable to ours, since they use different GPU implementations and, in some cases, different experimental protocol to do the measurements.

\section{Experimental Results}
\label{sec:results}

This section presents the experimental results obtained in this work.
Section~\ref{sec:ablation} presents preliminaries ablation studies conducted on a subset of the ImageNet dataset in order to investigate the effects of different channel selection strategies, image resolution, JPEG quality, JPEG chroma subsampling schemes and classification task difficulty.
Section~\ref{sec:channels_results} shows the effects our proposed methods for reducing the number of input channels;
Section~\ref{sec:lcrfa-results} shows the effects of removing the upsampling operation operation of the model;
Section~\ref{sec:layers_results} shows the results of our experiments reducing the number o layers.
Finally, Section~\ref{sec:time} presents a deeper analyses on the efficiency of the models, comparing inference time and FPS of the networks.

\subsection{Ablation Studies}
\label{sec:ablation}

Our ablation studies were conducted on a subset of the ImageNet dataset.
The initial experiments had the objective of validating our handcrafted approachs, the FBS technique, testing different coefficient ranges, image resolution, JPEG quality and chroma subsampling schemes.
For this, we evaluated our methods on the coarse grained classification difficulty of the subset.  
We started by testing different ranges of DCT coefficients to analyse which one contain the most relevant information.
For this, we selected 32 coefficients from each color channel with four different strategies: Selecting the lowest (coefficients 1 to 32), median (coefficients 17 to 48), highest (coefficients 33 to 64) and extremes frequencies (coefficients 1 to 16 and 49 to 64). The results can be seen in Table~\ref{tab:accranges}.

\begin{table}[!htb]
    \centering
    \caption{Comparison of accuracy for using DCT coefficients from different frequency ranges as input to the network on the coarse grained classification task from the ImageNet subset.}
    \begin{tabular}{lc}
        \hline
        \hline
        \textbf{Approach} & \textbf{Accuracy} \\
        \hline
            Lowest frequencies   & 94.53 \\
            Median frequencies   & 93.27 \\
            Highest frequencies  & 90.67 \\
            Extreme frequencies  & 94.12 \\
        \hline
        \hline
    \end{tabular}
    \label{tab:accranges}
\end{table}

The best performing strategy was selecting the lowest frequency coefficients, the second and third best were selecting the extreme and median frequencies, respectively, both including some coefficients from the lowest frequencies, while selecting only the highest frequencies was the worse strategy.
These results indicate that our initial hypothesis, that higher frequency information has little visual effect on the image, was correct.
For this reason, selecting the lowest frequency coefficients is the best strategy, and was used in all other experiments with the FBS method.


In our next experiment, we analysed the effects of different spatial resolution, as seen in Table~\ref{tab:resolution}, where values inside parentheses are the number of input channels of each network. 
For this, we resized all the images to have their smallest side with 256, 128, 64, and 32 pixels and used the crop sizes of 224$\times$224, 112$\times$112, 56$\times$56, and 28$\times$28, respectively.
As it can be seen, the reduction in the spatial resolution yielded a significant impact on the classification accuracy of all the networks, although the DCT-based ones were more affected, significantly reducing the accuracy. This may have occurred due to the low initial spatial resolution of the DCT, that is 8 times lower than the RGB.

\begin{table}[!htb]
    \small
    \centering
    \caption{Comparison of classification accuracy (\%) for the original ResNet-50 with RGB inputs, the Upsampling-RFA and its variants using our FBS technique on the coarse classification task of the ImageNet subset with multiple image resolutions.}
    \scriptsize
    \begin{tabular}{lcccc}
        \hline
        \hline
        \multirow{2}{*}{\textbf{Approach}} & \multicolumn{4}{c}{\textbf{Image Resolution}} \\
        \cline{2-5}
        & \textit{32} & \textit{64} & \textit{128} & \textit{256}\\
        \hline
        ResNet-50 (3x1)~\cite{CVPR_2016_He}              & 81.82 & 90.39 & 94.56 & 96.49  \\
        Upsampling-RFA~(3x64)~\cite{NIPS_2018_Gueguen}   & 72.72 & 82.06 & 90.32 & 94.15  \\
        Upsampling-RFA + FBS~(3x32)                      & 71.83 & 82.22 & 90.78 & 94.53 \\
        Upsampling-RFA + FBS~(3x16)                      & 70.35 & 81.35 & 90.16 & 93.92 \\
        \hline
        \hline
    \end{tabular}
    \label{tab:resolution}
\end{table}

We also analyzed the effect of reducing the quality setting used to encode JPEG images, as can be observed in Table~\ref{tab:quality}. All the networks were robust to this parameter, yielding small drops in accuracy, lower than 1\% for the worse quality setting. 

\begin{table}[!htb]
    \small
    \centering
    \caption{Comparison of classification accuracy (\%) for the original ResNet-50 with RGB inputs, the Upsampling-RFA and its variants using our FBS technique on the coarse classification task of the ImageNet subset and images with different JPEG qualities.}
    \scriptsize
    \begin{tabular}{lcccc}
        \hline
        \hline
        \multirow{2}{*}{\textbf{Approach}} & \multicolumn{4}{c}{\textbf{JPEG Quality}} \\
        \cline{2-5}
        & \textit{25} & \textit{50} & \textit{75} & \textit{100}\\
        \hline
        ResNet-50~(3x1)~\cite{CVPR_2016_He}            & 95.78 & 95.98 & 96.09 & 96.49\\
        Upsampling-RFA~(3x64)~\cite{NIPS_2018_Gueguen} & 93.84 & 94.02 & 94.50 & 94.15\\
        Upsampling-RFA + FBS~(3x32)                    & 93.63 & 93.97 & 94.20 & 94.53\\
        Upsampling-RFA + FBS~(3x16)                    & 92.69 & 93.26 & 93.66 & 93.92\\
        \hline
        \hline
    \end{tabular}
    \label{tab:quality}
\end{table}

Then we tested the impact of some of the most commonly used JPEG chroma subsampling schemes in the DCT models, being 4:4:4, 4:2:2, and  4:2:0, where the resolution is reduced in half both horizontally and vertically. We can see the accuracy obtained by each chroma subsampling scheme in Table~\ref{tab:subsampling}.

\begin{table}[!htb]
    \small
    \centering
    \caption{Comparison of accuracy (\%) for the Upsampling-RFA and its variants using our FBS technique on the coarse classification task of the ImageNet subset with multiple JPEG subsampling schemes.}
    \scriptsize
    \begin{tabular}{lccc}
        \hline
        \hline
        \multirow{2}{*}{\textbf{Approach}} & \multicolumn{3}{c}{\textbf{Chroma Subsampling}} \\
        \cline{2-4}
        & \textit{4:2:0} & \textit{4:2:2} & \textit{4:4:4} \\
        \hline
        Upsampling-RFA~(3x64)~\cite{NIPS_2018_Gueguen}  & 93.92 & 94.41 & 94.72  \\
        Upsampling-RFA + FBS~(3x32)                     & 94.53 & 94.88 & 95.16  \\
        Upsampling-RFA + FBS~(3x16)                     & 94.15 & 94.50 & 95.23  \\
        \hline
        \hline
    \end{tabular}
    \label{tab:subsampling}
\end{table}

The subsampling schemes with more chroma data available, obtained better results in most cases, but the increase in performance was very small for all models, 0.8\% at best between the subsampling with the smallest amount of data, the 4:2:0 and the one with the most data, 4:4:4.
Although the increase in performance was small, the size of the dataset in storage increased considerably.
The ImageNet subset we used in this experiment occupies 20.34 Gb of storage space encoded in JPEG with the 4:2:0 chroma subsampling scheme, 24.14 Gb with 4:2:2 and 31.34 Gb with 4:4:4.
This way, although the increase in accuracy is small, only up to 0.8\%, the increase in space occupied in disk is up to 54.08\%.
For this motive, we used the 4:2:0 chroma subsampling scheme in all the other experiments in this work, since it was the original encoding of the dataset, the fact that this scheme occupies the smallest amount of storage space and other subsampling schemes did not offer a considerable increase in performance.

In our last experiment with this subset, we considered both classification difficulties available, coarse and fine grained, and our proposed methods to reduce the number of input channels, handcrafted and data-driven, as can be seen in Table~\ref{tab:complexity1}

\begin{table}[!htb]
    \small
    \centering
    \caption{Comparison of classification accuracy (\%) for the ImageNet subset with different classification difficulty levels with 
    the original ResNet-50 with RGB inputs, the Upsampling-RFA and its variants using our FBS and data-driven channel reduction techniques on the ImageNet subset.}
    \scriptsize
    \begin{tabular}{lcc}
        \hline
        \hline

        \multirow{3}{*}{\textbf{Approach}} & \multicolumn{2}{c}{\textbf{Classification Task}} \\
        \cline{2-3}
        & \textit{Fine} & \textit{Coarse} \\
        & \textit{(211 Classes)} & \textit{(12 Classes)} \\
        
        \hline
        ResNet-50~(3x1)~\cite{CVPR_2016_He}            & 76.28   & 96.49 \\
        Upsampling-RFA~(3x64)~\cite{NIPS_2018_Gueguen} & 70.28   & 94.15 \\
        Upsampling-RFA + FBS~(3x32)                    & 69.79   & 94.53 \\
        Upsampling-RFA + FBS~(3x16)                    & 68.12   & 93.92 \\
        
        Upsampling-RFA + LP~(1x64)                     & 70.08   & 93.17\\
        Upsampling-RFA + LA~(1x64)                     & 69.15   & 94.23\\
        Upsampling-RFA + CCPP~(1x64)                   & 70.09   & 94.85\\ 
        \hline
        \hline
    \end{tabular}
    \label{tab:complexity1}
\end{table}

All networks obtained a higher classification accuracy  on the coarse grained task than on the fine grained task.
This result can be attributed to fact that the fine grained has a higher amount of classes than the coarse grained, being  a more challenging task.
For both tasks, the RGB-based network (ResNet-50) had better classification accuracy than the DCT-based ones (Upsampling-RFA variants). 
In the fine-grained task, the second best network was the Upsampling-RFA~\cite{NIPS_2018_Gueguen}, which obtained a result 6\% lower than the RGB. 

In the coarse-grained task, all our data-driven techniques were able to outperform the classification accuracy of the Upsampling-RFA, while at the same time, having lower computational complexity and amount of parameters.
Among them, CCPP was the best, with accuracy only 1.64\% lower than the RGB baseline.
Our handcrafted approach, when using 32 coefficients (Upsampling-RFA + FBS~(3x32)) was also able to obtain better classification accuracy than the Upsampling-RFA, being 1.96\% lower than the RGB.

\subsection{Effects of Reducing the Number of Input Channels}
\label{sec:channels_results}

We evaluated the computational complexity, amount of parameters, frames per second and accuracy of both our handcrafted and data-driven approaches for reducing the amount of input channels from the Upsampning-RFA.
Experiments were run ImageNet dataset, as it can be seen in Table~\ref{tab:accimagenet}.

\begin{table}[!htb]
    \centering
    \caption{Comparison of computational complexity (GFLOPs), number of parameters, frames per second on inference and classification accuracy for the original ResNet-50 with RGB inputs and the Upsampling-RFA variants using DCT as input for image classification on the ImageNet dataset.}
    \scriptsize
    \begin{tabular}{lcccc}
        \hline
        \hline
        
        \textbf{Approach} & \textbf{GFLOPs} & \textbf{Params} & \textbf{FPS} & \textbf{Accuracy} \\
        \hline
            \multicolumn{4}{l}{\underline{Gueguen~et~al.~\cite{NIPS_2018_Gueguen} training:}} \\
            
            ResNet-50~(3x1)             & 3.86 & 25.6M & 208 & 75.78 \\
            
            Upsampling-RFA~(3x64)       & 5.40 & 28.4M & 266 & 75.94 \\
            
            LC-RFA~(3x64)               & 5.11 & 27.4M & 267 & 75.92 \\
            
            Deconvolution-RFA~(3x64)    & 5.39 & 28.4M & 268 & 76.06 \\
            
            \hline
            
            \multicolumn{4}{l}{\underline{Deguerre~et~al.~\cite{deguerre2021object} training:}}\\
            
            ResNet-50~(3x1)           & 3.86 & 25.6M & 324 & 74.73 \\
            
            LC-RFA~(3x64)             & 5.11 & 27.4M & 318 & 74.82 \\
            
            LC-RFA~Y~(1x64)           & 5.14 & 27.6M & 329 & 73.25 \\
            
            Deconvolution-RFA~(3x64)  & 5.39 & 28.4M & 319 & 74.55 \\
            
            \hline
            
            \multicolumn{4}{l}{\underline{Our training:}}\\
            
            ResNet-50~(3x1)              & 3.86 & 25.6M &  588 & 73.46 \\
            
            Upsampling-RFA~(3x64)        & 5.40 & 28.4M &  494 & 72.33 \\

            LC-RFA~(3x64)   & 5.11 & 27.4M & 510 & 71.67 \\
            
            Upsampling-RFA + FBS~(3x32)  & 3.68 & 26.2M & 616 & 70.22 \\
            Upsampling-RFA + FBS~(3x16)  & 3.18 & 25.6M & 645 & 67.03 \\
            
            Upsampling-RFA + LP~(1x64)   & 3.20 & 25.6M &  492 & 69.62\\
            Upsampling-RFA + LA~(1x64)   & 3.20 & 25.6M &  626 & 69.96\\
            Upsampling-RFA + CCPP~(1x64) & 3.20 & 25.6M &  639 & 69.73\\
            
        \hline
        \hline
    \end{tabular}
    \label{tab:accimagenet}
\end{table}

The three models with the best classification accuracy were, respectively, the Deconvolution-RFA, Upsampling-RFA and LC-RFA trained by Gueguen~et~al.~\cite{NIPS_2018_Gueguen}.
All these models use DCT coefficients as inputs, and were able to achieve better classification accuracy than the RGB baselines.
The training  of these networks made by Deguerre~et~al.~\cite{deguerre2021object} and us were able to obtain classification accuracy similar to the RGB, but did not surpassed it.
In general, when comparing the same models trained by different authors, the best accuracy were obtained by Gueguen~et~al.~\cite{NIPS_2018_Gueguen}, followed by Deguerre~et~al.~\cite{deguerre2021object} and us, showing that small differences in procedures and training variance can affect the performance of these models.

The FPS obtained were also different for each work, since different hardware, batch size and other configurations were used.
In Gueguen~et~al.~\cite{NIPS_2018_Gueguen} experiments, the DCT models were able to obtain higher FPS than the RGB baseline, while for Deguerre~et~al.~\cite{deguerre2021object}, the LC-RFA and Deconcolution-RFA obtained lower FPS than the RGB model by a small margin, 6 and 5 frames, respectively, achieving a similar inference speed.

For this reasons, in the remainder of this work, when multiple training results for the same network were available, we used ours to keep a fair comparison.

Among the DCT models trained by Gueguen~et~al.~\cite{NIPS_2018_Gueguen} and Deguerre~et~al.~\cite{deguerre2021object}, the one with the best classification accuracy only increased it by 0.36\% compared to their RGB baseline, while the one with lowest complexity increased it by 32.38\% and had at least 1.8M more parameters.
Also Deguerre~et~al.~\cite{deguerre2021object} LC-RFA and Deconvolution-RFA obtained similar but smaller FPS than the ResNet-50.
The results obtained by Deguerre~et~al.~\cite{deguerre2021object} and us shows that the modifications made to accommodate DCT coefficients made to the ResNet-50, creating the Upsampling-RFA and LC-RFA, increased the computational complexity and number of parameters of the network, making it slower than the original model.
In our experiments, similar to Deguerre~et~al.~\cite{deguerre2021object}, the baseline Upsampling-RFA and LC-RFA obtained FPS lower than the RGB by 94 and 78 frames, respectively,
had a small drop in classification accuracy (around 1.13\% and 1.79\%) demanded an increase in computational complexity (of approximately 39.9\% and  32.38\%) and more trainable parameters (2.8M and 1.8M more).
Our Upsampling-RFA~+~FBS~(3x32), that uses 32 coefficients per color channel, obtained a classification accuracy 3.24\% lower and had 0.6M more parameters than our RGB baseline, but was able to reduce the complexity in 4.66\% and gain 28 FPS.
The Upsampling-RFA~+~FBS~(3x16) presented a greater drop in classification accuracy of 6.43\%, but also an even greater increases in efficiency, complexity was reduced in approximately 17.62\%, 57 FPS were gained and the amount of parameters was approximately equal to the RGB baseline.
All our data-driven strategies obtained similar classification accuracy, computational complexity, number of parameters and FPS.
Their amount of parameters were approximately the same as our RGB baseline, the complexity were approximately 17.1\% lower and there was a increase between 38 and 51 FPS.
LA performed slightly better, obtaining classification accuracy 3.5\% lower than the RGB baseline, followed closely by CCPP and LP, that were 3.73\% and 3.84\% lower, respectively.




Compared to the FBS techniques, our data-driven strategies yielded a similar classification accuracy and FPS to Upsampling-RFA + FBS~(3x32), while having a computational complexity and amount of parameters similar to Upsampling-RFA + FBS~(3x16).

\subsection{Effects of Avoiding the Chroma Upsampling}
\label{sec:lcrfa-results}

After testing our strategies for reducing the number of input channels on the Upsampling-RFA, we attempted to adapt them to the LC-RFA in order to avoid the cost of applying the upsampling operation to the chroma components of the DCT, decreasing even further the preprocessing cost.

We focused on our data-driven strategies, since they obtained a good balance between accuracy and computational cost.
Among them, we chose the CCPP because it is commonly applied in CNNs in order to obtain channel-wise projections of the feature maps, like in depthwise separable convolutions~\cite{chollet2017xception}.
Two experiments were made, the first one, CCPP~(1x32$|$1x64), used two CCPP layer, one to reduce the amount of channels from Y to n=32 and the other to reduce the concatenated Cb and Cr to 2n=64 channels.
The second one, CCPP~(1x16$|$1x32), reduce the amount of Y channels to n=16 and CbCr to 2n=32 channels.
Table~\ref{tab:accimagenet-lcrfa} shows the computational complexity, amount of parameters, accuracy on the ImageNet dataset and FPS of the LC-RFA and our improved variants.
Both our LC-RFA~+~CCPP~(1x32$|$1x64) and LC-RFA~+~CCPP~(1x16$|$1x32) obtained similar results. The one with CCPP~(1x32$|$1x64) got slightly higher accuracy, while the one with CCPP~(1x16$|$1x32) got slightly lower number of parameters, computational complexity and higher FPS.
The similarity in the results is due to the reduction in the number of channels in the LC-RFA with CCPP only affecting the first convolutional layer of the network from each stream, since the remainder of the model is kept with the same amount of input and output channels as the original ResNet-50.
This is different from our FBS strategies, that scale the amount of filter from the stages 2 and 3 of the network according to the amount of input channels.
Comparing to the ResNet-50 for RGB images, the LC-RFA + CCPP~(3x32) reduced the computational complexity in 18.65\%, had 0.9M less parameters and obtained FPS 28 frames higher, while the accuracy was 2.42\% lower.
Similarly, for LC-RFA + CCPP~(1x16$|$1x32), the computational complexity was 18.91\% lower, also had 0.9M less parameters and obtained 36 more FPS, while there was a loss in accuracy of 3.62\%.
As we can see, the results obtained for these method were similar to the Upsampling-RFA~+~CCPP, having slightly  better computational cost while the amount of parameters and FPS was slightly worse.
The CCPP~(1x16$|$1x32) version gained 0.11\% accuracy over the Upsampling-RFA~+~CCPP, being very similar, while the  CCPP~(1x16$|$1x32) gained 1.31\% accuracy, being our DCT model with the highest accuracy by a very small margin. 
This was expected, since similar strategies were used to reduce the cost of both of these networks.
This way, the LC-RFA models combined with the CCPP strategies can offer a good balance between accuracy and computational cost.

\begin{table}[!htb]
    \centering
    \caption{Comparison of computational complexity (GFLOPs), number of parameters, frames per second on inference and classification accuracy for methods based on the Upsampling-RFA and methods based on the LC-RFA.}
    \scriptsize
    \begin{tabular}{lcccc}
        \hline
        \hline
        
        \textbf{Approach} & \textbf{GFLOPs} & \textbf{Params} & \textbf{FPS} & \textbf{Accuracy} \\
        \hline

            ResNet-50~(3x1)              & 3.86 & 25.6M &  588 & 73.46 \\
            
            Upsampling-RFA~(3x64)        & 5.40 & 28.4M &  494 & 72.33 \\

            LC-RFA~(3x64)                & 5.11 & 27.4M &  510 & 71.67 \\
                       
            Upsampling-RFA + CCPP~(1x64) & 3.20 & 25.6M & 639 & 69.73\\

            LC-RFA + CCPP~(1x32$|$1x64)  & 3.14 & 24.7M &  616 &  71.04 \\
            LC-RFA + CCPP~(1x16$|$1x32)  & 3.13 & 24.7M &  624 &  69.84 \\ 
        \hline
        \hline
    \end{tabular}
    \label{tab:accimagenet-lcrfa}
\end{table}

\subsection{Effects of Reducing the Number of layers}
\label{sec:layers_results}

In this section, we will present the results we obtained by our strategy to reduce the amount of layers from the Upsampling-RFA and LC-RFA, proposed in Sections~\ref{sec:layers}~and~\ref{sec:lcrfa-ccpp}.
For the Upsampling-RFA, when stages of the network are skipped, we decreased or increase the amount of inputs channels in order to match the amount of channels expected at the initial stage in which they are provided as input.
For this, we use the CCPP method, since the results for all the strategies presented in Section~\ref{sec:channels} were similar.
This strategy was chosen because it is commonly applied in CNNs in order to obtain channel-wise projections~\cite{chollet2017xception}.
Table~\ref{tab:complexity} presents the computational complexity and number of parameters of our Upsampling-RFA~+~CCPP skipping different stages.
As it can be seen, skipping the first and second stages was beneficial, reducing the computational complexity and number of parameters of the network.
However, as more stages were skipped, although the number of parameters is decreased, the computational complexity is greatly increased.
This is due to the fact that by removing the earlier stages of the network, we are feeding inputs with higher resolutions to the network, since those early stages would reduce the resolution of the input.
For this reason and due to the decreased strides at the early blocks of the initial stage, the convolution operations have their cost increased though the network.

\begin{table}[!htb]
    \small
    \centering
    \caption{Comparison of computational complexity (GFLOPS) and number of parameters for our Upsampling-RFA + CCPP when skipping different stages.}
    \scriptsize
    \begin{tabular}{lcc}
        \hline
        \hline
        \textbf{Approach} & \textbf{GFLOPs} & \textbf{Params} \\
        \hline
        Skip the first stage                             &  3.20 & 25.6M \\
        Skip the first and second stages                 &  2.86 & 25.1M \\
        Skip the first, second, and third stages         &  8.26 & 23.9M \\
        Skip the first, second, third, and fourth stages & 10.76 & 15.8M \\
        \hline
        \hline
    \end{tabular}
    \label{tab:complexity}
\end{table}

Due to those motives, skipping of the first and second stages is the only setting considered in the next experiments, since only it saves the computational costs of the network.   

Table~\ref{tab:results} compares the computational complexity, number of parameters, frames per second and accuracy between our trained models presented previously and our strategy skip the first and second stages and use CCPP to accommodate the DCT inputs.
Skipping the first and second stages of the Upsampling-RFA benefited not only computational costs, but also the accuracy.
It achieved the best FPS and fourth best classification accuracy among the DCT-based networks.
It lost in classification accuracy only to the Upsampling-RFA and LC-RFA~\cite{NIPS_2018_Gueguen}, whose FPS,  computational complexity and number of parameters worse than the original ResNet-50, and to the LC-RFA~+~CCPP~(3x32) that had higher computational complexity and lower FPS.
Compared to the original ResNet-50 the classification accuracy was only 2.97\% lower, while there was a similar amount of parameters, a reduction of 25.91\% in computational complexity and a gain of 183 FPS.  

\begin{table}[!htb]
    \small
    \centering
    \caption{Comparison of computational complexity (GFLOPS), number of parameters, frames per second on inference and classification accuracy on the ImageNet dataset for the original ResNet-50 with RGB, networks designed for DCT, and our strategies for reducing the number of input channels and layers.}
    \scriptsize
    \begin{tabular}{lcccc}
        \hline
        \hline
        \textbf{Approach} & \textbf{GFLOPs} & \textbf{Params} & \textbf{FPS} & \textbf{Accuracy}\\
        \hline
        ResNet-50~(3x1)              & 3.86 & 25.6M &  588 & 73.46 \\
        
        Upsampling-RFA~(3x64)        & 5.40 & 28.4M &  494 & 72.33 \\

        LC-RFA~(3x64)   & 5.11 & 27.4M &  510 & 72.75 \\
            
        Upsampling-RFA + FBS~(3x32)  & 3.68 & 26.2M &  616 & 70.22 \\
        Upsampling-RFA + FBS~(3x16)  & 3.18 & 25.6M &  645 & 67.03\\
        
        \hline
        Upsampling-RFA + CCPP~(1x64) & 3.20 & 25.6M &  639 & 69.73\\

        LC-RFA + CCPP~(1x32$|$1x64) & 3.14 & 24.7M &  616 &  71.04 \\
        LC-RFA + CCPP~(1x16$|$1x32) & 3.13 & 24.7M &  624 &  69.84 \\ 
        
        \hline
        
        Upsampling-RFA + CCPP + skipping &  \multirow{2}{*}{2.86} & \multirow{2}{*}{25.1M} & \multirow{2}{*}{771} & \multirow{2}{*}{70.49}\\
        1$^{\mathrm{st}}$ and 2$^{\mathrm{nd}}$ stages~(1x128)\\

        LC-RFA + CCPP + skipping &  \multirow{2}{*}{2.91} & \multirow{2}{*}{24.4M} & \multirow{2}{*}{735} & \multirow{2}{*}{70.04}\\
        1$^{\mathrm{st}}$ and 2$^{\mathrm{nd}}$ stages~(1x32$|$1x64)\\

        LC-RFA + CCPP + skipping &  \multirow{2}{*}{2.90} & \multirow{2}{*}{24.4M} & \multirow{2}{*}{727} & \multirow{2}{*}{69.14}\\
            1$^{\mathrm{st}}$ and 2$^{\mathrm{nd}}$ stages~(1x16$|$1x32)\\ 
        \hline
        \hline
    \end{tabular}
    \label{tab:results}
\end{table}

Similar results were obtained when skipping the first and second stage of the LC-RFA models in terms of computational cost, but their classification accuracy were slightly lower than their counterpart without skip.
The model with CCPP~(1x32$|$1x64) obtained computational complexity 24.61\% lower than the original ResNet-50, 1.5M less parameters, the lowest amount of all network shown in this work, 147 more FPS while having a loss in accuracy of 3.42\%.
The model with CCPP~(1x16$|$1x32) had the same amount of parameters than the CCPP~(1x32$|$1x64), computational complexity 24.87\% lower than the ResNet-50, 139 more FPS and accuracy 4.32\% lower.
The small differences in FPS can be justified by the variance on the experiments.
Overall, these models obtained the lowest amount of parameters from all models we tested, but were outperformed in computational complexity, FPS and accuracy by the Upsampling-RFA~+~CCPP skipping the same stages.

\subsection{Analysing the Efficiency of the Models}
\label{sec:time}

Table~\ref{tab:times_and_fps} presents the results of our experiment to measure inference time and FPS.
As can be seen, the data preprocessing time is lower than the network time, although it still has a considerable influence, being approximately 16.48\% from the total inference time for the Upsampling-RFA, making the total FPS be 82 frames lower than the network FPS. 
It is important to emphasize that such results are hardware and software dependent and thus may change according to the experimental setup.
The base RGB model, the ResNet-50, was the one with the highest data preprocessing time and overall prediction time, since it needs to fully decode the images before feeding them to the network.
By using DCT coefficients that can be obtained by partial decoding of the images, the Upsampling-RFA and LC-RFA were able to reduce the data preprocesing time when compared to the ResNet-50, but since the modifications made leaded to a higher network time, there were drops in FPS for both models, since the increase in the network time was more influential.
The LC-RFA obtained considerably lower data preprocessing time than the Upsampling-RFA, this is due to the fact that no upsampling operation is applied to the Cb and Cr components, avoiding this cost, and that the the resolution of these components are halved in relation to the Y component, reducing the cost of operations to be applied.

\begin{table*}[!htb]
    \small
    \centering
    \caption{Comparison of inference time and frames per second for the original ResNet-50 with RGB, the Upsampling-RFA and LC-RFA designed for DCT, and our strategies for reducing the number of input channels and layers. The inference times are averaged for 10 runs over 25 batches of size 8, and \textpm indicates the standard deviation.}
    \scriptsize
\begin{tabular}{lccccc}
        \hline
        \hline
        \multirow{2}{*}{\textbf{Approach}} & \multicolumn{3}{c}{\textbf{Inference Times (ms)}} & \multicolumn{2}{c}{\textbf{FPS}} \\
        
        \cline{2-4}
        \cline{5-6}
        
        & \textbf{Preprocessing} & \textbf{Network} & \textbf{Total} & \textbf{Network} & \textbf{Total}  \\
        \hline
        ResNet-50~(3x1)              & 85.848 \textpm 2.640 & 339.704 \textpm 5.428 & 425.615 \textpm 6.689 & 588 & 469 \\

        \hline
        
        Upsampling-RFA~(3x64)        & 79.820 \textpm 2.015 & 404.396 \textpm 2.471 & 484.268 \textpm 3.396 & 494 & 412 \\
        
        Upsampling-RFA + FBS~(3x32)  & 53.492 \textpm 2.222 & 324.331 \textpm 6.872 & 377.870 \textpm 8.585 & 616 & 529 \\
        
        Upsampling-RFA + FBS~(3x16)  & 39.698 \textpm 1.834 & 310.072 \textpm 9.765 & 349.815 \textpm 10.662 & 645 & 571 \\
        
        \hline
        Upsampling-RFA + LP~(1x64)   & 80.240 \textpm 2.082 & 406.354 \textpm 6.505 & 486.651 \textpm 7.925   & 492 & 410 \\ 
        
        Upsampling-RFA + LA~(1x64)   & 82.484 \textpm 2.953 & 319.416 \textpm 5.744 & 401.955 \textpm 7.667   & 629 & 497 \\ 

        Upsampling-RFA + CCPP~(1x64) & 83.099 \textpm 2.718 & 312.987 \textpm 18.200 & 396.134 \textpm 20.276 & 639 & 504 \\
        
        Upsampling-RFA + CCPP + skipping 1$^{\mathrm{st}}$ and 2$^{\mathrm{nd}}$ stages~(1x128) & 83.232 \textpm 3.599 & 259.335 \textpm 7.847  & 342.624 \textpm 9.908 & 771 & 583 \\

        \hline
        LC-RFA~(3x64)          & 69.379 \textpm 5.261 & 391.433 \textpm 4.049  & 460.857 \textpm 8.454  & 510 & 433 \\

        LC-RFA + CCPP~(1x32$|$1x64)   & 69.688 \textpm 2.516 & 324.484 \textpm 17.518 & 394.219 \textpm 19.123 & 616 & 507 \\

        LC-RFA + CCPP~(1x16$|$1x32)  & 68.247 \textpm 3.707 & 320.320 \textpm 18.981 & 388.610 \textpm 22.341 & 624 & 514 \\

        LC-RFA + CCPP + skipping 1$^{\mathrm{st}}$ and 2$^{\mathrm{nd}}$ stages~(1x32$|$1x64) & 69.770 \textpm 2.338 & 271.874 \textpm 16.434 & 341.687 \textpm 18.625 & 735 & 585 \\

        LC-RFA + CCPP + skipping 1$^{\mathrm{st}}$ and 2$^{\mathrm{nd}}$ stages~(1x16$|$1x32) & 70.366 \textpm 3.820 & 275.031 \textpm 15.022 & 345.441 \textpm 18.151 & 727 & 578 \\

        \hline
        \hline
    \end{tabular}
    \label{tab:times_and_fps}
\end{table*}

By using the FBS technique on the Upsampling-RFA, the data preprocessing time was improved even further, together with a reduction in network time, generating big improvements on the FPS of the networks over the Upsampling-RFA, LC-RFA and RGB baseline.
Our data-driven methods for reducing the amount of input channels also greatly improved the network time, but obtained similar data preprocessing time to the the model they were applied to, being the Upsampling-RFA, or LC-RFA.
This indicates that the main source of reduction in preprocessing time for the FBS methods is the discard of the unused DCT coefficients, reducing the cost of applying transformations on the data (like transposing, slicing and concatenation of tensors) and the time to load the data into the GPU.
Our data-driven methods reduce the amount of channels after the aforementioned operations are executed, since they need to be incorporated to the network to be trained together, and for this reason, only obtain the gain in data preprocessing time for partially decoding the images.
But they were able to get FPS similar to FBS~(3x32), since their network time was also similar.
Skipping stages of the networks obtained similar gains in data preprocessing time as the Upsampling-RFA since it also use our data-driven strategy to reduce the amount of input channels, but they were the strategies with best network time among the tested models.
The results we obtained here indicates that methods for improving the time of data preprocessing, like using network that work directly on compressed data and our FBS and data-driven techniques to reduce the amount of input channels are promising, since they reduce data preprocessing time  as well as network time, being able to greatly speed-up the models.

\begin{figure}[!htb]
    \centering
    \includegraphics[width=0.9\columnwidth]{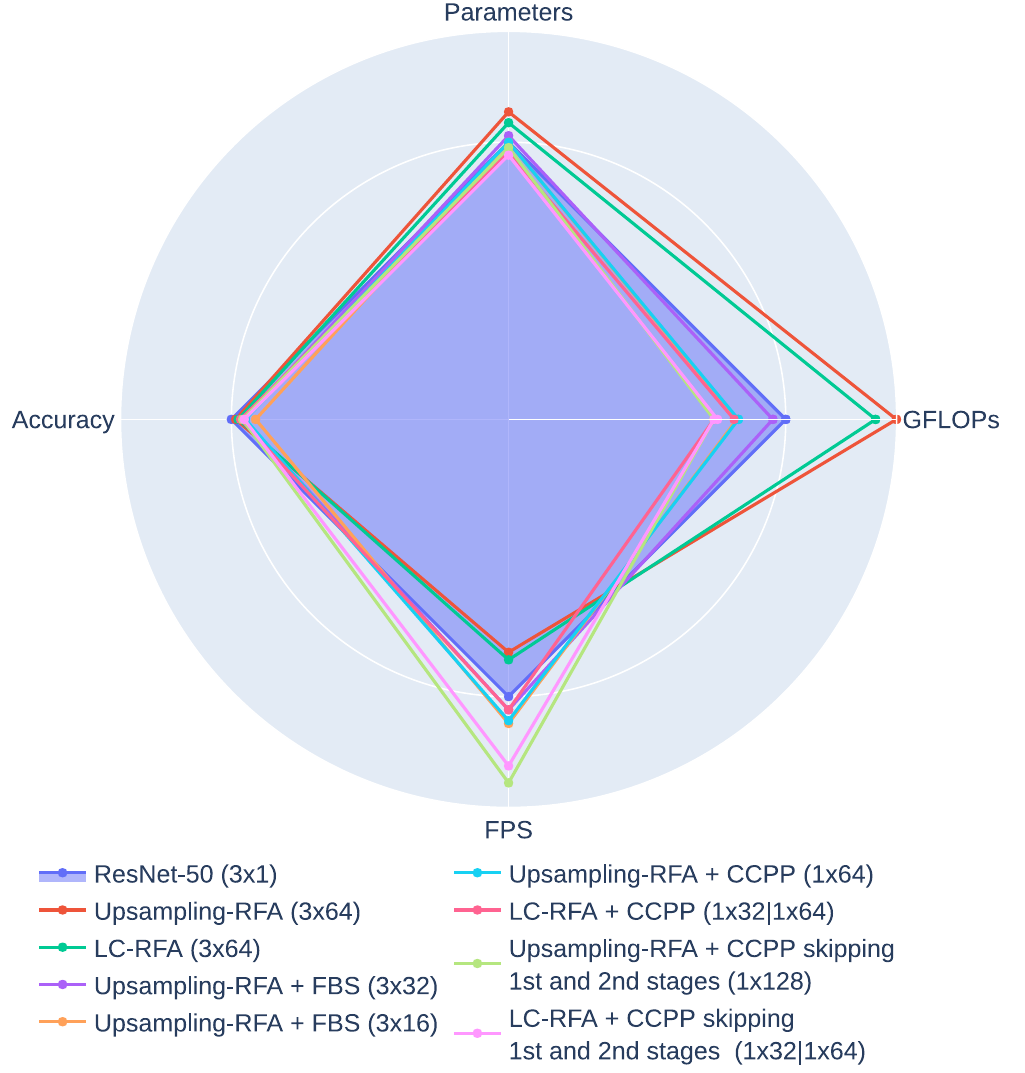}
    \caption{Comparison of computational complexity, number of parameters, frames per second and classification accuracy from different networks for the ImageNet dataset. Values are shown in proportion to the ResNet-50.}
    \label{fig:imagenet}
\end{figure}

Figure~\ref{fig:imagenet} compares the computational complexity, number of parameters, frames per second, and classification accuracy for the different networks evaluated on the ImageNet dataset.
As it can be seen, although all networks have similar classification accuracy, the Upsampling-RFA and LC-RFA significantly increased the computational complexity and number of parameters from the original ResNet-50, only leading to a small speed-up in relation to the RGB baseline.
The use of the FBS technique in the Upsampling-RFA was able to reduce the computational complexity and the number of parameters while keeping a similar classification accuracy and greatly increasing the FPS of the network.
Our data-driven approaches combined with the Upsampling-RFA obtained similar classification accuracy to the FBS~(3x32), FPS higher than the Upsampling-RFA but lower than the FBS~(3x32) and a similar amount of parameters and computational complexity to the FBS~(3x16).
When applied to the LC-RFA, similar computational complexity and FPS were obtained while the accuracy was slightly higher, being the DCT model with the third best, behind only to the Upsampling-RFA and LC-RFA that have considerably higher computational cost.
By using a data-driven technique to reduce the amount of channels and skipping the 1$^{\mathrm{st}}$ and 2$^{\mathrm{nd}}$ stages, we obtained the DCT model with the lowest computational complexity, amount of parameters and highest FPS, while keeping similar accuracy.

\section{Conclusions}
\label{sec:conclusions}

In this paper, we presented a study on CNNs designed to operate directly on frequency domain data, learning with DCT coefficients rather than RGB pixels.
These information are readily available in the compressed representation of images, saving the high computational load of fully decoding the data and greatly speeding up the processing time, which is currently a big bottleneck of deep learning, being crucial for environments with limit computational resources, like edge devices and embedded systems.
The starting point of our work was the state-of-the-art models proposed by Gueguen et al.~\cite{NIPS_2018_Gueguen}, the Upsampling-RFA and LC-RFA, which are modified versions of the ResNet-50 architecture~\cite{CVPR_2016_He}.
Despite the speed-up obtained by partially decoding JPEG images, their architectural changes raised the computation complexity and the number of parameters of the network.
To alleviate these drawbacks, we propose a Frequency Band Selection (FBS) technique to select the most relevant DCT coefficients before feeding them to the network.

Initially, we made ablation studies in a subset of ImageNet, where we showed that selecting the lowest frequency coefficients of the DCT was the best between all tested strategies, DCT models are sensitive to variations in image resolution and robust to changes in JPEG quality and chroma subsampling schemes.
The classification accuracy for the methods that use the FBS technique in the frequency domain can be similar to state-of-the-art methods, but its inference speed can be increased in up to 57 FPS.
Although the FBS technique is effective, relevant information from the DCT inputs are discarded.
For this reason, we explored smart strategies to reduce the computational complexity without discarding useful information.
Our results showed that learning how to combine all DCT inputs in a data-driven fashion  was able to reduce computational complexity and the amount of parameters, reaching an classification accuracy and FPS similar to the FBS with 32 coefficients for each color channel. 
Also, we found that skipping some stages of the network is beneficial, leading to model with a good balance between classification accuracy, FPS and amount of parameters, proving to be an efficient strategy.
The results we obtained in this work indicates that designing model to work in directly with data from the compressed domain is promising, being able to obtain similar accuracies while greatly improving the computational cost of these models.
As future work, we intend to evaluate other smart strategies to reduce computational complexity of the network.
We also plan to extend our approach for video classification using 3D network architectures, like Res3D~\cite{ARXIV_2017_Tran} or I3D~\cite{CVPR_2017_Carreira}.
As well as exploring auto-encoders in order make learn how to reduce the amount of input channels, following strategies like Chen~et~al.~\cite{chen2018learning}.
Finally, we intend to evaluate our approach in large-scale datasets, like Kinetics~\cite{ARXIV_2017_Kay}, and in other applications.

\section*{Acknowledgment}
This research was supported by the FAPESP-Microsoft Research Virtual Institute (grant~2017/25908-6) and the Brazilian National Council for Scientific and Technological Development - CNPq (grant~314868/2020-8).
We gratefully acknowledge the support of NVIDIA Corporation with the donation of the Titan Xp GPU used for this research.

\bibliographystyle{IEEEtran}
\bibliography{bibliography.bib}

\end{document}